\documentclass[preprint]{elsarticle}
\pdfoutput=1
\usepackage{tipa,booktabs}  
\usepackage{helvet}         
\usepackage{courier}        
\usepackage{type1cm}        
\usepackage{makeidx}         
\usepackage{graphicx}        
\usepackage{multicol}        
\usepackage[bottom]{footmisc}
\usepackage[utf8]{inputenc}
\usepackage{lipsum}
\makeatletter
\def\ps@pprintTitle{%
 \let\@oddhead\@empty
 \let\@evenhead\@empty
 \def\@oddfoot{}%
 \let\@evenfoot\@oddfoot}
\makeatother

\begin{document}
\begin{frontmatter}

\title{Hierarchical Representation of Prosody for Statistical Speech Synthesis}

\cortext[cor1]{Corresponding author}

\author[mv,as]{Antti Suni}
\author[da1,da2,mv]{Daniel Aalto}
\author[mv]{Martti Vainio\corref{cor1}}

\address[mv]{Institute of Behavioural Sciences, University of Helsinki}
\address[as]{Department of Signal Processing and Acoustics, Aalto University}
\address[da1]{Communication Sciences and Disorders, Faculty of Rehabilitation Sciences, University of Alberta}
\address[da2]{Institute for Reconstructive Sciences in Medicine (iRSM), Misericordia Hospital, Edmonton}

\begin{abstract}

Prominences and boundaries are the essential constituents of prosodic structure in speech.  They provide for means to chunk the speech stream into linguistically relevant units by providing them with relative saliences and demarcating them within coherent utterance structures.  Prominences and boundaries have both been widely used in both basic research on prosody as well as in text-to-speech synthesis.  However, there are no representation schemes that would provide for both estimating and modelling them in a unified fashion. Here we present an unsupervised unified account for estimating and representing prosodic prominences and boundaries using a scale-space analysis based on continuous wavelet transform.  The methods are evaluated and compared to earlier work using the Boston University Radio News corpus.  The results show that the proposed method is comparable with the best published supervised annotation methods. 

\end{abstract}

\begin{keyword}
phonetics \sep prosody \sep speech synthesis \sep wavelets
\end{keyword}

\end{frontmatter}


\section{Introduction}

Two of the most primary features of speech prosody have to do with chunking speech into linguistically relevant units above the segment and the relative salience of the given units; that is, boundaries and prominences, respectively.  These two aspects are present in every utterance and are central to any representation of speech prosody; moreover, they give rise to a hierarchy.  Ideally they would be represented with a uniform methodology that would take into account both the production and the perceptual aspects of the speech signals.  Such a system would be beneficial to both basic speech research and speech technology, especially speech synthesis.  On the other hand, to be useful for data oriented research and technology, the system should strive towards being unsupervised as opposed to annotation systems that rely on humans.  Ideally the system would still behave in a human-like fashion, while avoiding the subjectiveness and variability caused by the blend of top-down and bottom-up influences involved in the interpretation of linguistic speech signals.

In this paper we present a hierarchical, time-frequency scale-space analysis of prosodic signals (e.g., fundamental frequency, energy, duration) based on the continuous wavelet transform (CWT). The presented algorithms can be used to analyse and annotate speech signals in an entirely unsupervised fashion.  The work stems from a need to annotate speech corpora automatically for text-to-speech synthesis (TTS) \cite{s4a} and the subject matter is mainly examined from that point of view. However, the presented representations should be of interest to anyone working on speech prosody.

Wavelets extend the classical Fourier theory by replacing a fixed window with a family of scaled windows resulting in scalograms, resembling the spectrogram commonly used for analysing speech signals. The most interesting aspect of wavelet analysis with respect to speech is that it resembles the perceptual hierarchical structures related to prosody. In scalograms speech sounds, syllables, (phonological) words, and phrases can be localised precisely in both time and frequency (scale). This would be considerably more difficult to achieve with traditional spectrograms. Furthermore, the wavelets give natural means to discretise and operationalise the continuous prosodic signals.

\begin{figure}[t]
\begin{center}
\includegraphics[width=\linewidth]{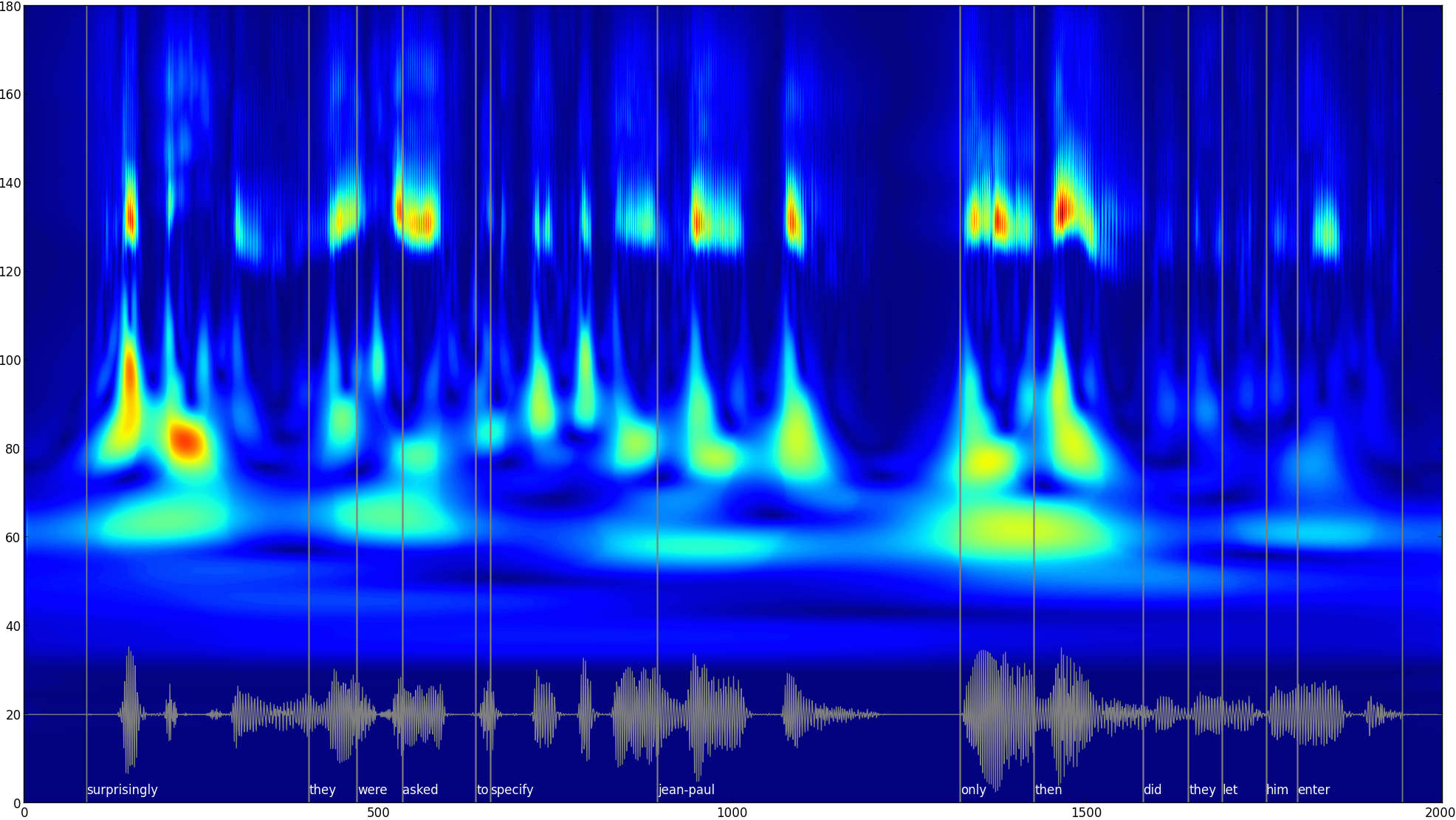}
\caption{A continuous wavelet transform based on the signal envelope of an English utterance showing the hierarchical structure of speech.  The lower pane shows the waveform and the upper pane the CWT.  In the CWT figure, the wavelet scale diminishes towards the top of the figure.  The lower parts show the syllables as well as prosodic words (see text for more detail).}
\end{center}
\label{fig:hierarchy}
\end{figure}

Figure \ref{fig:hierarchy} depicts the hierarchical nature of speech as captured in a time-frequency scale-space by CWT of the signal envelope of a typical English utterance. The upper part contains the formant structure (which is not visible due to the rectification of the signal) as well as the fundamental frequency in terms of separate glottal pulses.  Underneath the $f_0$ scale are the separate speech segments followed by (prominent) syllables, as well as prosodic words.  The lower part including the syllables and prosodic words depicts the suprasegmental and prosodic structure which has typically not been represented in e.g., the mel-frequency cepstral coefficient (MFCC) based features in both ASR and TTS. 

Spoken language is organised hierarchically both structurally and phonetically: words belong to phrases and are built up from syllables which are further divisible into phonemes which stand for the actual speech sounds when the structures are realised as speech.  This has many non-obvious effects on the speech signal that need to be modelled.  The assumption of hierarchical structure combined with new deep learning algorithms has lead to recent breakthroughs in automatic speech recognition \cite{deng2013machine}.  In synthesis the assumption has played a key role for considerably longer. The prosodic hierarchy has been central in TTS since 1970's \cite{hertz1982text,hertz1985delta} and most current systems are based on some kind of a hierarchical utterance structure.  Few systems go above single utterances (which typically represent sentence in written form), but some take the paragraph sized units as a basis of production \cite{HLT}.

The hierarchical utterance structure serves as a basis for modelling the prosody, e.g., speech melody, timing, and stress structure of the synthetic speech.  Controlling prosody in synthesis has been based on a number of different theoretical approaches stemming from both phonological considerations as well as phonetic ones.  The phonologically based ones stem from the so called Autosegmental Metrical theory \cite{goldsmith1990autosegmental} which is based on the three-dimensional phonology developed in \cite{halle1978metrical,halle1980three} as noted in \cite{klatt1987review}.  These models are sequential in nature and the hierarchical structure is only implicated in certain features of the models.  The more phonetically oriented hierarchical models are based on the assumption that prosody -- especially intonation -- is truly hierarchical in a super-positional and parallel fashion.  

Actual models capturing the superpositional nature of intonation were first proposed in \cite{ohman1967word} by Öhman, whose model was further developed by Fujisaki \cite{fujisaki,fujisaki_hirose84} as a so called command-response model which assumes two separate types of articulatory commands; accents associated with stressed syllables superposed on phrases with their own commands.  The accent commands produce faster changes which are superposed on a slowly varying phrase contours.  Several superpositional models with a varying degree of levels have been proposed since Fujisaki \cite{bailly2005sfc,anumanchipalli2011statistical,kochanski2000stem,kochanski2003prosody}.  Superpositional models attempt to capture both the chunking of speech into phrases as well the highlighting of words within an utterance.  Typically smaller scale changes, caused by e.g., the modulation of the airflow (and consequently the $f_0$) by the closing of the vocal tract during certain consonants, are not modelled. 

Prominence is a functional phonological phenomenon that signals syntagmatic relations of units within an utterance by highlighting some parts of the speech signal while attenuating others.  Thus, for instance, some of syllables within a word stand out as stressed \cite{eriksson-syllable}.  At the level of words prominence relations can signal how important the speaker considers each word in relation to others in the same utterance.  These often information based relations range from simple phrasal structures (e.g., prime minister, yellow car) to relating utterances to each other in discourse as in the case of contrastive focus (e.g., "Where did you leave your car? No, we WALKED here."). Although prominence probably functions in a continuous fashion, it is relatively easily categorised in e.g, four levels where the first level stands for words that are not stressed in any fashion prosodically to moderately stressed and stressed and finally words that are emphasised (as the word WALKED in the example above).  These four categories are fairly easily and consistently labeled even by non-expert listeners \cite{cole2010signal,Vainio2009,arnold2012obtaining}. In sum, prominence functions to structure utterances in a hierarchical fashion that directs the listener's attention in a way which enables the understanding of the message in an optimal manner. However, prominent units -- be they words or syllables -- do not by themselves demarcate the speech signal but are accompanied by boundaries that chunk the prominent and non-prominent units into larger ones: syllables to (phonological) words, words to phrases, and so forth. Prominence and boundary estimation have been treated as separate problems stemming from different sources in the speech signals.   

As functional -- rather than a formal -- prosodic phenomena prominences and boundaries lend themselves optimally to statistical modelling. The actual signalling of prosody in terms of speech parameters is extremely complex and context sensitive -- the form follows function in a complex fashion.  As  one-dimensional features, prominence and boundary values provide for a means to reduce the representational complexity of speech annotations in an advantageous way. In a synthesis system it occurs at a juncture that is relevant in terms of both representations and data scarcity.  The complex feature set that is known to effect the prosody of speech can be reduced to a few categories or a single continuum from dozens of context sensitive features, such as e.g, part-of-speech and whatever can be computed from the input text. Taken this way, both word prominence and boundaries can be viewed as abstract phonological functions that impact the phonetic realisation of the speech signal predictably and that can show considerable phonetic variation in its manifestation.  They are essential constituents of the utterance structure, whereas features like part-of-speech or information content (which are typically used for predicting prosody) are not. 

Word prominence has been shown to work well in TTS for a number of languages, even for English which has been characterised as a so called intonation language \cite{becker2006rule,Suni2010,Suni2011,Suni2012,badino2012towards}. In principle English should require a more detailed modelling scheme with explicit knowledge about the intonational forms. The perceived prominence of a given word in an utterance is a product of many separate sources of information; mostly signal based although other linguistic factors can modulate the perception \cite{jop,Vainio2009}.  Typically a prominent word is accompanied with a $f_0$ movement, the stressed syllable is longer in duration, and its intensity is higher.  However, estimating prominences automatically is not straight-forward and a multitude of differenct estimation algorithms have been suggested (see Section \ref{chap:experiment} for more detail).

Statistical speech synthesis requires relatively little data as opposed to unit-selection based synthesis.  However, labelling even small amounts of speech -- especially by experts -- is prohibitively time consuming.  In order to be practicable the labelling of any feature in the synthesis training data should be preferably attainable with automatic and unsupervised means.  

In what follows we present recently developed methods for automatic prominence estimation based on CWT (Section \ref{chap:methods}) which allow for fully automatic and unsupervised means to estimate both (word) prominences and boundary values from a hierarchical representation of speech (see \cite{Suni2013b,vainio2013continuous,vainioetal2015} for earlier work).  The main insight in this methodology is that both prominences and boundaries can be treated as arising from the same sources in the (prosodic) speech signals and estimated with exactly the same methods.  These methods, then, provide for a uniform representation for prosody that is useful in both speech synthesis and basic phonetic research. These representations are purely computational and thus objective.  It is -- however -- interesting to see how the proposed methods relate to annotations provided by humans as well as earlier attempts at the problem (Section \ref{chap:experiment}).

\section{Methods} \label{chap:methods}
 
Wavelets are used in a great variety of applications
for effectively compressing and denoising signals,
to represent 
the hierarchical properties of multidimensional signals
like polychromatic visual patterns in image retrieval,
and to model optical signal processing of visual neural fields
 \cite{russ1995image,romeny2014}.
In speech and auditory research  there is also a long 
history going back to the 1970's \cite{zweig1976,altosaar1988multiple,yang1992,
ramachandran1995modern,ridge,reimann2011,giraud12}. 
A recent summary of wavelets in speech technology can be found in 
\cite{farouk2014application}.

\begin{figure*}[tph]
\centerline{\includegraphics[width=\linewidth]{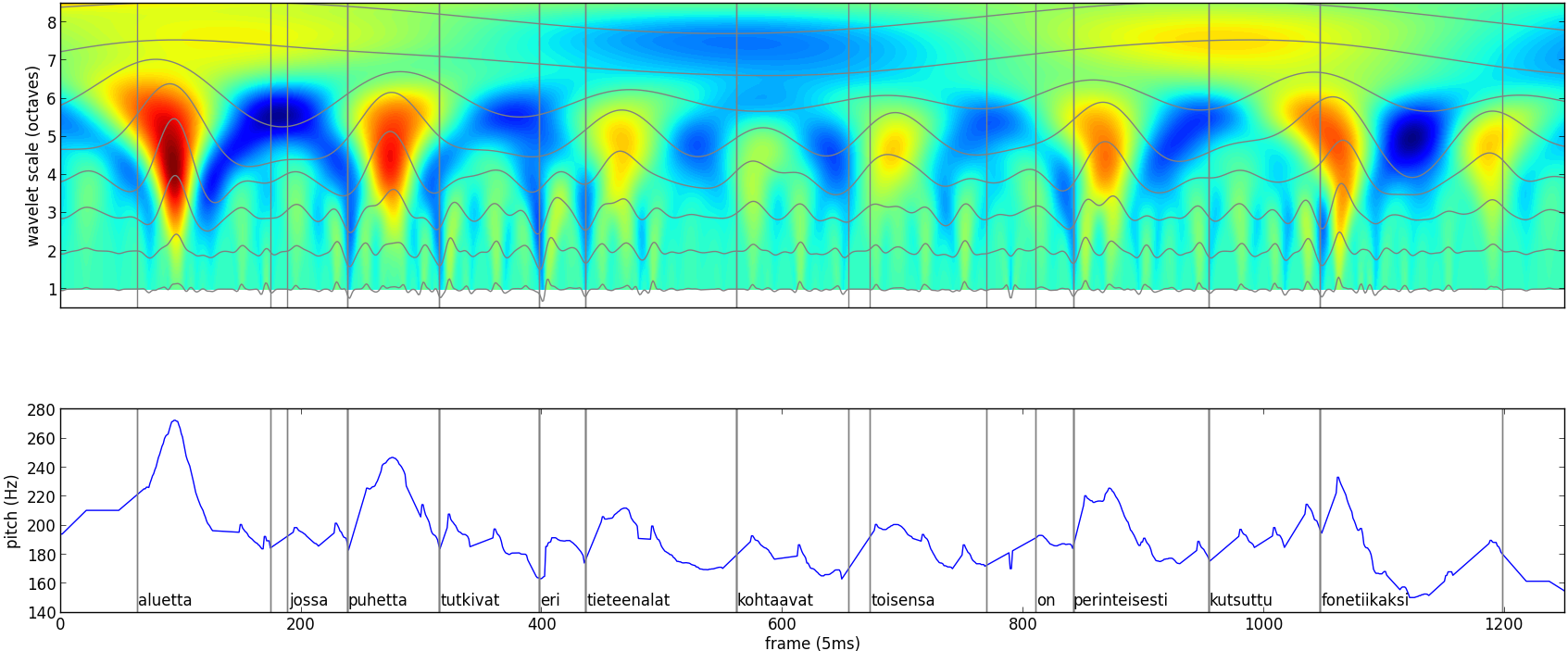}}
\caption{CWT of the $f_0$ contour of a Finnish utterance. The lower pane shows the (interpolated) contour itself as well as orthographic words (word boundaries are shown as vertical lines in both panes).  The upper pane shows the wavelet transform as well as eight separated scales (grey lines) ranging from segmentally influenced perturbation or microprosody (lowest scale) to utterance level phrase structure (the highest level).}
\label{fig:Scales}
\end{figure*}

Figure \ref{fig:Scales} shows a CWT of the $f_0$ contour of a Finnish utterance ``Aluetta, jossa puhetta tutkivat eri tieteenalat kohtaavat toisensa on perinteisesti kutsuttu fonetiikaksi", (The area where the sciences interested in speech meet each other has been traditionally called phonetics.). The lower pane shows the (interpolated) contour itself as well as orthographic words (word boundaries are shown as vertical lines in both panes).  The upper pane shows the wavelet transform as well as eight separated scales (grey lines) ranging from segmentally influenced perturbation or microprosody (lowest scale) to utterance level phrase structure (the highest level).  The potentially prominent peaks in the signal occurring during most content words are clearly visible in the scalogram.

The time-scale analysis allows for not only locating the relevant features in the signal but also estimating their relative salience, i.e., their prominence.  The relative prominences of the different words are visible as positive local extrema (red in Fig. \ref{fig:Scales}). There are several ways to estimate word prominences from a CWT. Suni et al. \cite{Suni2013b} and Vainio et al. \cite{vainio2013continuous} used amplitude of the word prosody scale which was chosen from a discrete set of scales with ratio 2 between ascending scales as the one with the number of local maxima as close to the number of words in the corpus as possible. A more sophisticated way is presented in \cite{vainioetal2015} where the lines of maximum amplitude (LoMA) in the wavelet image were used \cite{mallat1999,ridge,grossman1985}. This method was shown to be on par with human estimated prominence values (on a four degree scale).  However, the method still suffers from the fact that not all prominent words are identified and -- more importantly -- some words are estimated as prominent whereas they should be seen as non-prominent parts of either another phonological word or a phrase.

\begin{figure*}[t]
\centerline{\includegraphics[width=\linewidth]{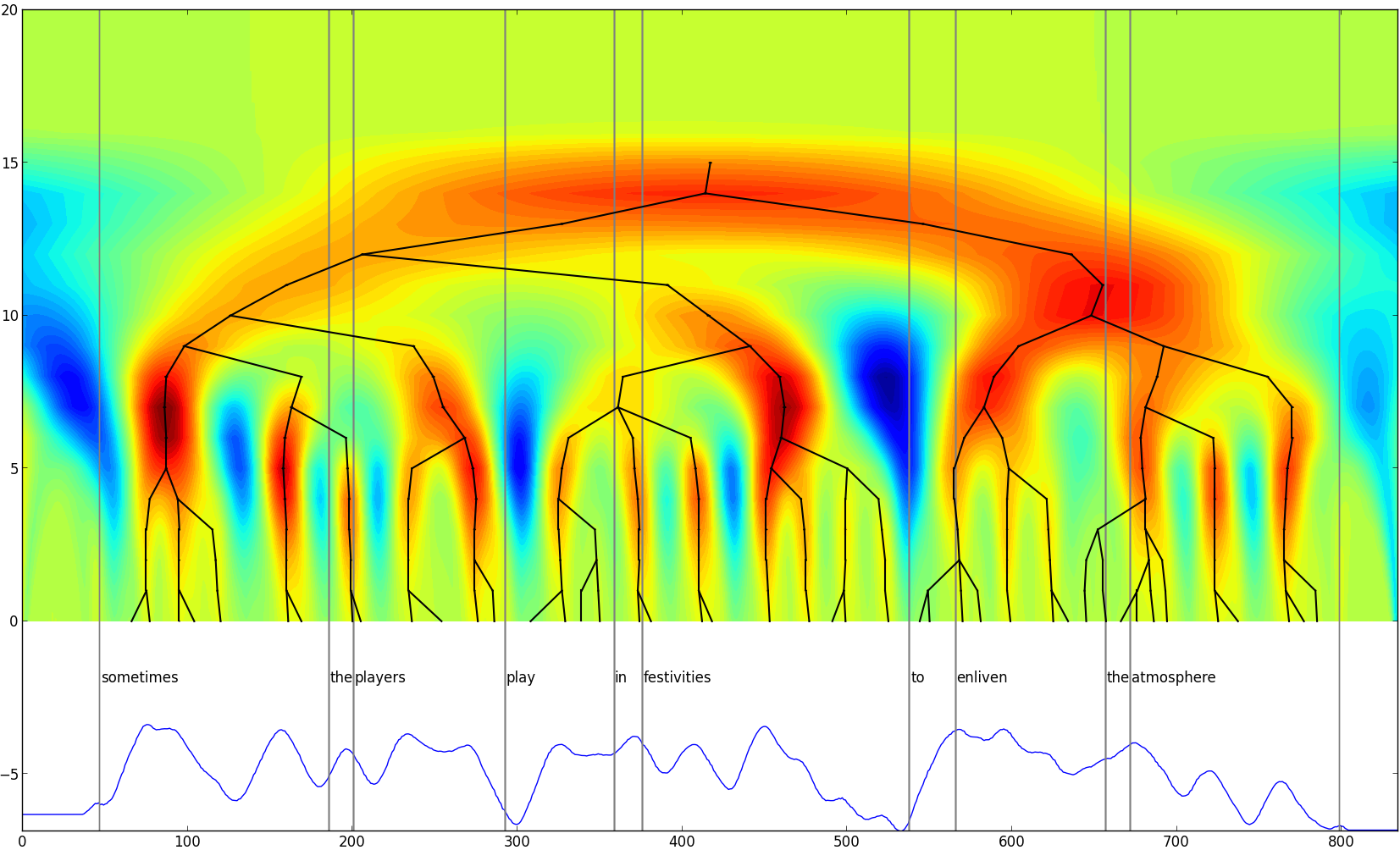}}
\caption{CWT based analysis of an utterance with hierarchical tree structure higlighted in black. Levels correspondig to syllables, prosodic words, and
phrases can be observed.}
\label{fig:tree}
\end{figure*}

Figure \ref{fig:tree} shows an $f_0$ contour of an English utterance (``Sometimes the players play in festivities to enliven the atmosphere.") analysed with CWT.  The analyses provide both an accurate measure for the locations of the prominent features in the signal as well as their magnitudes.  All in all, the CWT based analysis can be used for a fully automatic labelling of speech corpora for synthesis.  The synthesis, however, cannot produce a full CWT at run time; neither does it make sense to use the full transform for training.  That is, the CWT needs to be partitioned into meaningful scales for both training and producing the contours.

In earlier work, wavelets have been used in speech synthesis context mainly for parameter estimation \cite{kruschke,mishra,vansanten} but never as a full modelling paradigm. In the HMM based synthesis framework, decomposition of $f_0$ to its explicit hierarchical components during acoustic modelling has been investigated in \cite{iflytek,zen1}.  These approaches rely on exposing the training data to a level-dependent subset of questions for separating the layers of the prosodic hierarchy. The layers can then be modelled separately as individual streams \cite{iflytek}, or jointly with adaptive training methods \cite{zen1}.

In the current study we have extended the CWT based analysis by using two-dimensional tagging of prosodic structure; in addition to the LoMA based prominence we use a boundary value of each word in order to 1) better represent the hierarchical structure of the signal, and 2) to disambiguate those prominence estimates that are estimated to be similarly prominent by the LoMA estimation alone.  This brings the labelling system closer to the traditional tone-sequence models which have been widely used -- with varying rates of success -- in English TTS \cite{hirschberg1993pac,dutoit1997introduction,taylor2009text}. The boundary value for each word can be estimated by e.g, following the lines of minimum amplitude at word boundaries (blue areas in Figure \ref{fig:tree}).  The combination of word prominence and boundary values -- together with the traditional text based utterance structure -- are enough to represent the sound structure of any utterance.  These utterance structures can be further modified by other functional features such as whether the utterance is a question or a statement by simply adding the feature to the top-level of the tree.

The above described scheme reduces the complexity of the symbolic representation of speech at a juncture that optimises the learning of the actual phonetic features derived from the speech events -- be they parameter tracks or something else, such as e.g., articulatory gestures.

In the remaining part of the section we describe the main steps for analysing and annotating prominences and boundaries in a fully automatic and unsupervised fashion using the CWT and LoMA on composite prosodic signal based on fundamental frequency, intensity, and timing.

\subsection{Wavelet decomposition}

The basis for the modeling of hierarchies in speech signals
is provided by continuous wavelet transform (CWT).
The continuous theory is explained in detail
by Daubechies and the theory is applied
to time series as by Torrence and Compo 
\cite{daubechies,torrencecompo}.
The CWT is a decomposition of a signal in
scales which can be summed up to yield the original signal
approximately. To define the transform, let $s$ be a
one-dimensional signal with real values and finite energy.
Given a scale $\sigma>0$ and a temporal translation $\tau$,
the continuous wavelet transform can be defined as
$Ws(\sigma,\tau) = \sigma^{-1/2}s*\psi_{\tau,\sigma}$
where $*$ denotes the convolution and
$\psi_{\tau,\sigma}$ is the Mexican hat mother wavelet translated by $\tau$
and dilated by $\sigma$. Although the Mexican hat mother wavelet has
infinite support, the values decay exponentially fast far away
from the origin and the mother wavelet effectively acts on
a support of seven units. 

The sampling rate of a digital signal determines the finest temporal
scales available for the analysis. In the statistical
speech synthesis context a 5~ms fixed window size is used
for acoustical parameters. Every real signal also has finite length
and the coarsest scales become obsolete. The onset and
offset of the signal can create artifacts propagating to the wavelet
image and here these effects are
counteracted by continuing the signal periodically.

The original signal $s$ can be reconstructed approximately
from the original signal using a finite number of wavelet
scales with
$$
s(t) \approx c \sum_{j=0}^Na^{-j/2}Ws(a_0a^j,t)
$$
where $a_0>0$ is the finest (smallest) scale, 
$a>1$ defines the spacing between chosen scales, $N>1$ is
the number of scales included,
and $c$ is a constant. 
Throughout this work, $a=\sqrt{2}$.

\subsection{Lines of maximum amplitude}

\begin{figure}[t]
\begin{center}
\includegraphics[width=\linewidth]{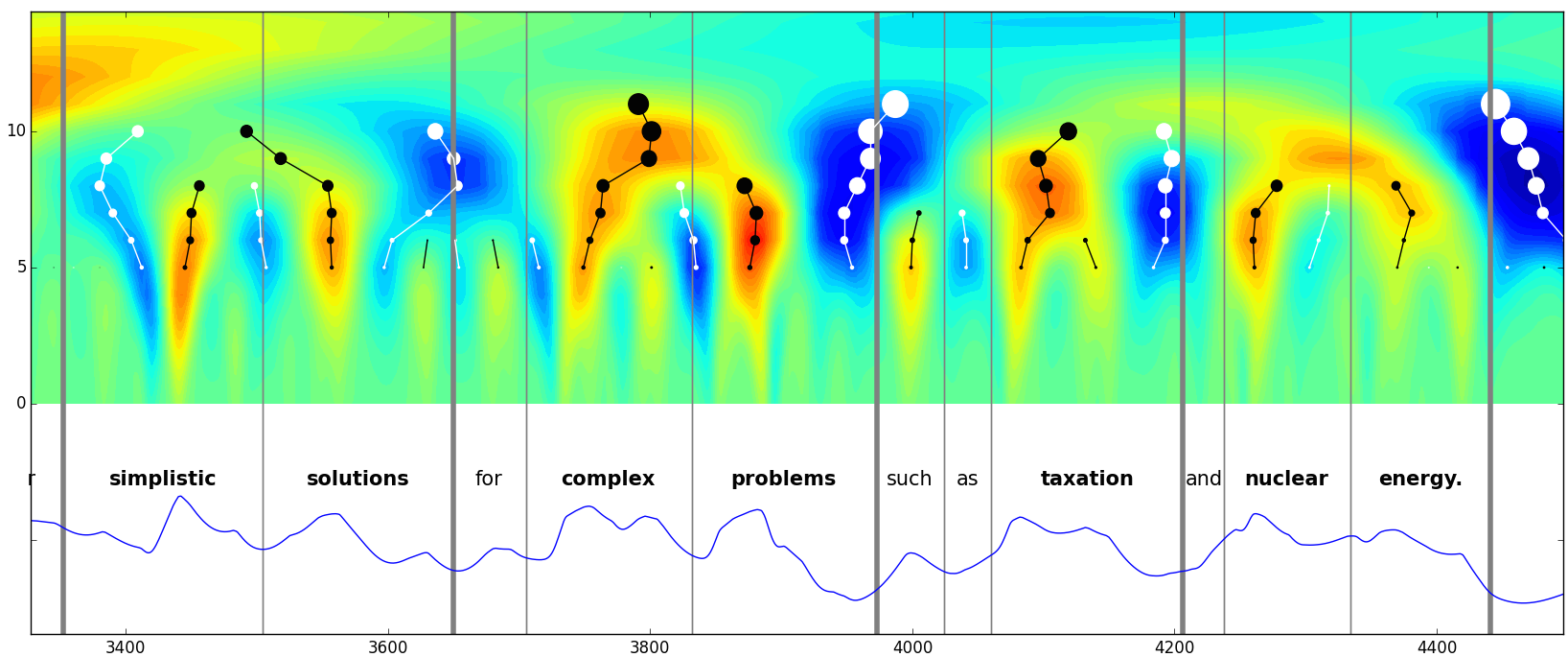}
\caption{A fragment of speech from BURNC analysed with CWT-LoMA with combined prosodic signal. 
Maxima lines are drawn in black and minima lines in white, with point size representing cumulative strength.
Annotated prosodic boundaries are marked with vertical lines and accented words with boldface type.}
\end{center}
\label{fig:loma}
\end{figure}

The Mexican hat mother wavelet belongs to a family of Gaussian wavelets. These wavelets seem to give a suitable compromise between temporal and frequency selectivity in the time-frequency representation of the prosodic signals. Importantly, the Gaussian wavelets give trees that allow for full reconstruction of the original signal. Visually, the trees look stable and consistent with Mexican hat mother wavelets.

Instead of a full tree representation of the prosody (as depicted in the Fig. 3.), a reduced tree representation is used here. Towards this purpose, lines of maximum amplitude (LoMA) are defined recursively by connecting local maxima across scales. First, let $t_{1,0}$, $t_{2,0}$,$\dots$, $t_{M_W,0}$ be the time points where the local maxima occurred in the finest scale ($\sigma=a_0$) in descending order, $Ws(a_0,t_{1,0})\geq \dots \leq  Ws(a_0,t_{M_W,0})$. Then the point $t_{i,0}$, $i=1,\dots,M_W$ is connected to the nearest local maximum (the mother candidate) to the right at the scale $a_0a$ if the derivative along the scale at $t_{i,0}$ is positive, the distance to the mother candidate is at most $200$~ms, and the mother candidate was not connected to a child earlier. If the derivative was negative, the search was done to the left. For consecutive levels, the ordering is based on the cumulative weighted sum of the local maximum together with its descendants: for a local maximum in $t_{i,j}$, $j>0$, at level $a_0a^j$, with descendants in $t_{i_0,0},\dots,t_{i_j,j}$ at levels $a_0,\dots,a_0a^j$ respectively, the cumulative weighted sum is 

$$ Ws(a_0,t_{i_0,0})+\dots+\log(j+1)a^{-j/2}Ws(a_0a^j,t_{i_j,j}).$$

Without the logarithmic term in the above sum, the formula resembles a lot the reconstruction of the original signal. Since the local maxima often are close to each other, the logarithmic term plays a crucial role in giving more weight to the higher levels of hierarchy. Observe that the number of local maxima decrease with increasing scales, every local maximum has at most one parent, and every parent has exactly one child. Finally, the points connected as children and parents form lines of maximum amplitude (LoMA) and the {\em strength} of such a line is the weighted sum of all the elements included in the line. 

The lines of minimum amplitude (LomA) of a signal $s$ are defined as the lines of maximum amplitude of $-s$. The positive and negative lines are then used for estimating prominence and boundary magnitudes, respectively. An example of LoMA analysis is shown in Figure 4.

\subsection{Preprocessing of the signals}

The acoustic signal reflects the physiological control actions behind speech communication. Emphasised words are often louder, higher pitch, and longer as a result of more production effort, higher fundamental frequency, and prolonged duration. For analysing the acoustic patterns, the abrupt changes in $f_0$ or gain, due to e.g. closures in the vocal tract during stops, create strong hierarchical structures in the wavelet image that might not be part of the auditory {\em gestalt} \cite{barnes2011}. Because of the more continuous underlying articulatory gestures and because of the seemingly more continuous percepts, the acoustic signals are ``filled in'' for the portions where signal cannot be found (for $f_0$) or where it is very weak (gain). In addition, a continuous (with respect to the time) representation for duration is derived. Although inspired by the physiology of vocal and auditory apparatuses, the aim of these transformations is not to model these systems but to make the algorithm more comparable to the other phenomenological approaches to describe the key prosodic patterns.
 
\begin{figure}[t]
\centerline{\includegraphics[width=\linewidth]{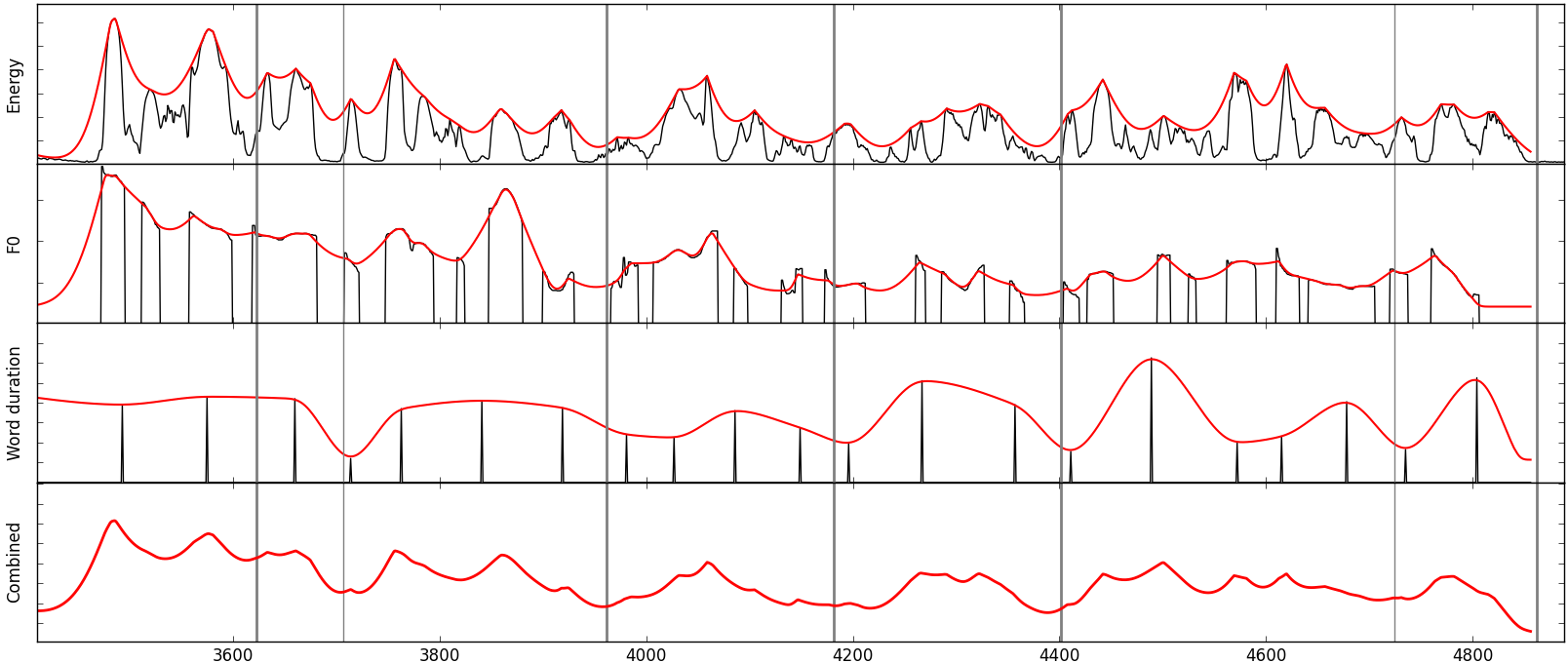}}
\caption{Prosodic parameters used in LoMA analysis extracted from BURNC. Raw parameters are drawn in gray and Interpolated final parameters are shown in red.
Combined prosodic signal is shown in the bottom. Gray vertical lines represent manually annotated prosodic boundaries.
}
\label{fig:pre}
\end{figure}

\subsubsection{Intensity}

Intensity variations in the speech signal are primarily
caused by (deliberate and random) fluctuations of
subglottal pressure and the degree of hyperarticulation (especially
in fricatives). As a proxy to the articulatory effort,
the gain of the acoustical signal
is transformed by iteratively interpolating the silent gaps.

Let $\phi$ be the Gaussian kernel
and $g$ the original gain signal (i.e. a logarithm of the amplitude).
A family of scaling functions, $\{\phi_i\}_i$ is obtained
by dilating and scaling $\phi$ with constants 
$\lambda_i = w_{\max}^{(i-n)/n}w_{\min}^{-i/n}$,
$i=0,1,2,\dots, n$,
where $w_{\max}$ is the maximum smoothing window size,
$w_{\min}$ is the minimum window,
and $n$ is the size of the family.
The $g$ is recursively smoothed. 
For $i=0$,
a pointwise maximum is taken by
$g_0 = \max \{g,g*\phi_0\}$
where $*$ denotes the convolution.
For $i>0$,
$g_i = \max \{g,g_{i-1}*\phi_{i}\}$.
This results in the preprocessed gain $g = g_n$ shown in the
top pane of Figure 5.

\subsubsection{Fundamental frequency}

The auditory pitch of the voiced sounds is closely related to the lowest eigen resonances of the vocal folds. However, during unvoiced speech segments, the association between the acoustic signal and the eigen resonances of the vocal folds break apart. Importantly, even during the silent periods there are control actions to the vocal folds that impact the $f_0$ once the vibration its reinitiated either by adducting the vocal folds or by restoring the airflow through vocal tract. In addition to the internal state of the larynx, the frequency of the glottal pulsing is influenced by the subglottal pressure. Not surprisingly then, the $f_0$ and intensity are strongly correlated. To estimate the state of the $f_0$ control during unvoiced portions, an algorithm is proposed where the surface $f_0$ values are left unchanged for the voiced passages and the underlying state of the vocal folds is estimated by interpolation for unvoiced passages.

The gap filling for the unvoiced portions of fundamental frequency signal $s$ is similar to that for the gain. First, the signal is decomposed in voiced and unvoiced portions by defining the set $V$ of time points where the speech signal is voiced.

In practice, the voicedness of a time point is defined using the GlottHMM \cite{raitio} analysis which applies low-frequency energy and zero-crossings thresholds for voicing decision. Then, using the same smoothing family as before, 
the smoothed $s$ is defined iteratively: for $i=0$,
$s_0 = s\chi_{V} + \max \{s,s* \phi_0 \}\chi_{V^C}$
where $\chi_A$ is the characteristic function of a set $A$
and $A^C$ denotes the complement of the set $A$.
The analogous recursive formula then is
$$
s_i = s\chi_{V} + \max \{s,s_{i-1}*\phi_{i} \} \chi_{V^C}
$$
resulting in the preprocessed fundamental frequency. Finally, to remove perturbation around gaps, the iterated signal $s_n$ is smoothed using the same iterated maximisation algorithm as for the gain.

To find suitable parameters in the above algorithms, two test utterances were used. These values were used:
$w_{\max} = 100$~ms, $w_{\min}=1$, for both gain and $f_0$;
$n=100$ for
for gain, $n=200$ for $f_0$; and for final smoothing of $f_0$
$w_{\max} = 25$~ms and $n = 50$.

Observe that the repeated convolutions and maximums do not let the signals grow in an unlimited way. Instead, every point converges and the resulting (maximal) function has comparable energy to the original which can be seen by iterating a result of Hardy and Littlewood \cite{hardylittlewood}, (for modern approach, see Theorem 2.19 in \cite{mattila}).

\subsubsection{Duration}

The duration of a phonological unit varies as a function of its position within an utterance. For instance the speech rate often changes across boundaries and accented words are longer.  Due to a lack of signal based speech rate estimators, the duration signal has to be based on analytical linguistic units rather than the raw signal. To quantify the duration, a relation between acoustical (continuous) duration and a suitable discrete linguistic unit is needed. A natural candidate could be a syllable but here an orthographic word is chosen instead as the syllable boundaries might not be easy to derive from text without supervision. To apply the wavelet analysis to the duration, it is expanded to a continuous time dependent variable which ideally would reflect the local duration of the linguistic units. For the current experiment provided word alignments were used. The word boundaries, 
$x_0,x_1,\dots,x_{N_w}$,
where $N_w$ is the number of orthographic words within a given speech signal, and the associated durations
$d_i = x_i-x_{i-1}$, $i=1,\dots,N_w$, are computed.
These points $\{(x_{i-1}+d_i/2,d_i)\}$ are connected using cubic splines to yield a duration signal $d$ defined for every
time instant from $x_0$ to $x_{N_w}$
with the same sampling rate as for fundamental frequency and gain. When annotated pauses and breaths between words occurred, these were not taken into consideration, i.e. these gaps were not interpolated.

\subsection{Annotation}

The annotation of accentsand breaks (prominences and boundaries) is based on wavelet decompositionof the fundamental frequency, gain, and duration signals.These three acoustic signalswere normalized to have unit varianceand then summed to yield aprosodic signal $s$.The finest scale to be analysed was defined asbeing one octave below the rate of occurence ofortographic words. To normalize the speech rate, the finest scalewas selected for each utterance separatelythrough finding the word scale $a_W$which is the ratio of word count and utterance duration.

The finest scale was one octave below word scale, i.e. $a_0 = 1/2a_W$, and the coarsest scale was three octaves higher,
i.e. $8a_0$. If no LoMA took place during the word, the accent strength was set to zero.

The prosodic breaks manifest mostly on larger scales, sothe word scale was taken as the finest scale $a_0 = a_W$.The coarsest scale was again three octaves higher.Instead of local maxima and theassociated LoMA, the local minima were used as a basisfor the break annotation. To approximate the local speech rate,the time derivative of the continuous duration wasused instead of the continuous duration.Then, the combined signal of scaled $f_0$, gain, and duration derivativeweresubject to LomA analysis.

\section{Experimental Results} \label{chap:experiment}
%

As stated in the introduction, a solid method for prosody annotation would be very welcome in speech synthesis field, where recent development has concentrated on acoustic modelling side \cite{zen}. The motivation is especially strong in building speech synthesizers for  low-resourced languages,  where neither linguistically nor prosodically annotated corpora are available \cite{s4a}. In this chapter, we asses the utility of the proposed CWT-LoMA representation of prosody on the tasks of unsupervised annotation of prosodic prominences and boundaries. Although this hierarchical method does lend itself naturally to multi-level prosody annotation \cite{vainio2013continuous},  here, we restrict ourselves to binary detection task, in order to produce comparable results with previous studies. Furthermore, in TTS binary prosodic labels can be a useful simplification, as it better facilites text based prediction.

Previous work on unsupervised prosody annotation has focused on accent or prominence. For example, Ananthakrishnan \& Narayanan \cite{shankar2006} performed two-class unsupervised clustering on syllable level acoustic features combined with lexical and syntactic features, achieving accent detection accuracy of 78\% using Boston University Radio News Corpus (BURNC). In a similar vein, Mehrabani et al. \cite{mehrabani} annotated a corpus with four level prominence scale  by K-means clustering on foot-level acoustic features, achieving improved synthesis quality compared to a rule-based prominence model.  Using more analytic approach, Tambourini \cite{tamburini} derived a continuous prominence function, using expert knowledge to weight various acoustic correlates of prominence, achieving 80\% accuracy on syllable prominence detection on TIMIT corpus. Word prominence was annotated by Vainio \& Suni \cite{specom} with similar method, using  prosodic features generated by parametric synthesis build without prominence labels as a powerful normalizing method. An ambitious approach  was presented by Kalinli \&  Narayanan \cite{kalinli_interspeech}, extracting multi-scale auditory features insipired on the processing stages  in the human  auditory system, combined to an auditory salience map. They achieved prominent word detection accuracy of 78\% with F-score of 0.82 on BURNC, which , to our knowledge, is the best reported unsupervised result on this corpus to date.

Whereas text-based break prediction literature is abundant due to its importance in TTS, unsupervised acoustic boundary annotation has received less interest. 
This probably stems from the fact that both acoustic pauses, which can be obtained reliably by HMM forced alignment, and punctuation yield high baseline accuracy on major boundaries, and
for TTS purposes, this has been considered satisfactory.  For example in BURNC, intonational phrase boundaries can be predicted by silence alone with 88\% accuracy, though with only 45\% recall, and  traditional acoustic features offer little improvent over this trivial baseline \cite{rosenberg2009automatic}.   In terms of combining text and acoustic evidence, Ananthakrishnan \& Narayanan \cite{shankar2006}  obtained 81\% accuracy in combined intermediate and intonational boundary detection with two class k-means model.

\subsection{Corpus}

We perform the evaluation of our prominence and boundary detection method  on Boston Radio News corpus \cite{boston}, chosen 
for high quality prosodic labeling and comparability with several previous methods also evaluated on BURNC. The corpus consists of 
about two and a half hours of news stories read by 6 speakers with manual Tone and Break Index annotations. 
The ToBi labelling scheme was originally developed for transcribing speech melody \cite{tobi}, thus high (H), low (L) and complex accent types are employed 
(H*, L*, L*+H, L+H*, H+ !H*),  concerned with syllable level  shape and peak alignment. Prosodic boundaries are annotated with boundary tones 
(L-, H-,L – L\%, L – H\%, H – H\%, H – L\%),  again signalling the shape of melody. Break strength is annotated in the form of break indices ranging from zero (clitized) to four (intonational phrase boundary). For the boundary detection task,  we consider a word boundary as a prosodic boundary if the last syllable of a preceding word is marked with break index three (intermediate phrase break) or four (intonational phrase break). Prominence, on the other hand, has not been directly annotated and for the current experiment, we make a simplifying assumption that word is prominent if any of it's syllables carries an accent. These binary boundary and prominence categories are consistent with previous prosodic event detection studies \cite{ananthakrishnan2008, kalinli_journal}. Almost all of the annotated data were used for the experiment, totalling 442 stories or 29774 words.  Three stories from speaker f2b, used for setting values of free parameters were excluded as well as few cases were syllable and word alignments did not match. Word level break and prominence labels were derived by combining the provided, time aligned syllable and word labels. Manually corrected alignments were used when available.

\subsection{Features and Processing}

The proposed method was evaluated using standard prosodic features; $f_0$, energy and word duration, as well as all combinations of those. 
Raw $f_0$ and energy parameters were analyzed from 16 kHz speech signals with GlottHMM analysis-synthesis framework \cite{raitio} with five millisecond frame shift. The method uses Iterative-adaptive inverse filtering to separate the contributions of vocal tract and voice source, and performs  $f_0$  analysis on the source signal with autocorrelation method. Log energy is calculated from the whole signal.  Pitch range was set separately for male and female speakers, 70--300 Hz and 120--400 Hz, respectively. Obtained  $f_0$  and energy parameters were interpolated using peak preserving method and  word durations were transformed to continuous signals as described in section 2.3. Labeled pauses and breaths were not considered in the duration transfom on the y-axis. When evaluating the performance of combinations of prosodic features, the individual parameters were normalized utterance-wise to zero mean, unity variance, and summed prior to the wavelet analysis, with no weight adjustments, after which the composite prosodic signal was again normalized. 

The signal was then used as such in wavelet analysis, without any feature extraction step. Continuous wavelet transform was performed using the second derivative of gaussian (Mexican hat) wavelet,  with a half octave scale separation. Scale corresponding to word level was estimated individually for each paragraph in order to normalise speech rate differences. Lines of maximum and minimum amplitude were then estimated from the scalogram. 
Strongest peak LoMA of each word was assigned as the prominence value of the word and strongest valley LomA between each two word's strongest peak LoMA
as a boundary value. If word contained no peak LoMA, valley LomA was searched between the midpoints of adjacent words. Further, if either peak LoMA or valley LomA was not found, prominence or boundary value was set to zero respectively. To verify the utility of hierarchical modelling and rule out the possibility that improvements were achieved only due to feature engineering, 
we also calculated word maximum (to represent prominence) and minimum between midpoints of adjacent words (to represent boundary) from raw prosodic signal to be used as a baseline.
In order to compare the predicted continuous prominence and boundary values against manual labels, the values were converted to binary form by searching for an optimal value for separating the two classes in terms of classification accuracy, using 10\% of the manual labels. Although continuous values could be used as such in other applications, it might be argued that this step weakens our claim for unsupervision for the current task. Thus, for the best configurations, we also report results based on dividing the prominence and boundary distributions to two classes by unsupervised k-means clustering.

\subsection{Results}

We report results on CWT-LoMA analysis of $f_0$ (f0) energy (en) and duration (dur) and their combinations on prominence and boundary detection. The performance of gap-filling on energy is evaluated separately, and whenever the gap-filling improves the performance for prominence or boundary annotation, it is used for energy in combined features as well. Boundaries were defined as manual break indices of either 3 or 4; prominence if any syllable of a word carries an accent. Results are presented in word level, in terms of percentage of correct detections, i.e. accuracy, as well as precision, recall and F-score. As baselines, we report the majority class, predictions derieved from best combination signal without wavelet analysis, as well as current state-of-the-art unsupervised and supervised acoustic results. Note that these results are only roughly comparable, as there are minor differences in data selection. Results are presented in Table 1. Strictly unsupervised results using two class k-means clustering on the prediction distributions using all acoustic features were 84.0\% accuracy and 0.86 F-score for prominence and  85.5\%, 0.73 for boundary detection respectively.

\begin{table}[ht]
\label{tab:results}
\centering
\caption{Summary of results on BURNC with comparison to earlier experiments.  The bolded figures depict the best results both in current experiments and the literature. $^{1}$Kalinli \&  Narayanan \cite{kalinli_interspeech},$^{2}$Rosenberg \& Hirchberg \cite{Rosenberg2009},$^{3}$Ananthakrishnan \& al. \cite{shankar2006},$^{4}$Ananthakrishnan \& al. \cite{ananthakrishnan2008}.}
\begin{tabular}{ll}
\small
\begin{tabular}{l|llll}
\toprule

\textbf{}                    & \multicolumn{4}{l}{\textbf{Prominence Detection}} \\ \hline
\textit{feature}             & \textit{acc.\%}  & \textit{F-score}  & \textit{prec.} & \textit{rec.} \\ \midrule
\textit{f0}                  & 80.9                   & 0.82              & 0.84               & 0.81            \\ \hline
\textit{en}                  & 79.5                   & 0.83              & 0.77               & 0.89            \\ \hline
\textit{en\_interp.}    & 78.3                   & 0.81              & 0.77               & 0.86            \\ \hline
\textit{dur}                 & 79.5                   & 0.81              & 0.81               & 0.81            \\ \hline
\textit{f0\_en}              & 82.5                   & 0.85              & 0.81               & 0.88            \\ \hline
\textit{f0\_dur}             & 84.2                   & \textbf{0.86}     & 0.85               & 0.87            \\ \hline
\textit{en\_dur}             & 82.5                   & 0.84              & 0.82               & 0.86            \\ \hline
\textit{f0\_en\_dur}         & \textbf{84.6}          & \textbf{0.86}     & 0.84               & 0.90            \\ \hline
\textbf{Baselines}	& & & & \\ \hline
\textit{majority}            & 54.5                   &                   &                    &                 \\ \hline
\textit{f0\_en\_dur\_raw}    & 79.2                   & 0.81              & 0.82               & 0.80            \\ \hline
\textit{unsupervised}$^{1,2}$        & 78.1                   & 0.82              & 0.78               & 0.86            \\ \hline
\textit{acoustic sup.}$^{3,4}$ & 84.2                   & \textbf{0.86}     &                    &                 \\ \hline
\end{tabular}

\vline

\begin{tabular}{lllll}

\toprule
\multicolumn{4}{l}{\textbf{Boundary Detection}}             \\ \hline
\textit{acc.\%} & \textit{F-score} & \textit{prec.} & \textit{rec.} \\ \midrule
81.1                 & 0.56             & 0.79               & 0.44            \\ \hline
78.6                 & 0.54             & 0.68               & 0.45            \\ \hline
81.0                 & 0.56            & 0.79               & 0.44            \\ \hline
80.3                 & 0.64             & 0.66               & 0.61            \\ \hline
81.7                 & 0.59             & 0.80               & 0.47            \\ \hline
\textbf{85.7}                 & 0.72             & 0.79               & 0.67            \\ \hline
85.2                 & \textbf{0.73}    & 0.75               & 0.70            \\ \hline
\textbf{85.7}        & 0.72             & 0.80               & 0.65            \\ \hline
                     &                  &                    &                 \\ \hline
72.0                 &                  &                    &                 \\ \hline
82.1                 & 0.62             & 0.76               & 0.53            \\ \hline
81.1                 & 0.66             & 0.64               & 0.69            \\ \hline
84.6                &                  &                    &                 \\ \hline
\end{tabular}
\end{tabular}
\vline

\end{table}
 
Examining the results of individual acoustic features, we note similar perfomance for $f_0$ and energy in both tasks, and word duration not far behind. $f_0$ appears more
important for prominence detection, which is expected as reference labeling concerned pitch accents.  Filling the unvoiced gaps of energy signal helps in boundary detection,
but not in the accent detection task, perhaps due to syllable level features of the signal being smoothed too much. Interestingly, combining  $f_0$ and energy yields only modest improvement,
whereas combining either with duration provides substantial gain; accuracy increases approximately 3\%  in accent detection and 4\% in boundary detection. Though a naïve feature, 
word duration may capture both lengthening effects as well as lexical information, separating most of the function and content words, and disambiguating the alignment of LoMA.
Combining all features provides best results, but not by significant margin. Comparison of  the detection estimates from raw combined signal to ones provided by CWT-LoMA confirm
the importance of hierarchical modelling with solid advantage in both tasks.

Compared to previous methods, our results improve opon unsupervised state-of-the-art by a significant margin, and at least match the accuracy of acoustic-based supervised methods. 
The results are not far from performance of supervised methods using acoustic, lexical, and syntactic evidence, where reported accuracies for both word level prominence and boundary detection range from 84\% to 87\%
 \cite{ananthakrishnan2008, kalinli_journal}.

\section{Discussion and Conclusions} \label{chap:conclusion}

In contrast to most published work on speech prosody, the results here show that prosodic structure can -- and probably should -- be studied and represented in a unified framework comprising all relevant signal variables at the same time. For statistical speech synthesis we are now in a position where we can perform full annotation and modelling of prosody in a unified framework. Although, we have presented the methods in the service of speech synthesis, the results are interesting by themselves.  That is, they show that prominences and boundaries can be viewed as manifestations of the same underlying speech production process.  This has, of course, many theoretical implications.  As foremost is the fact that the suprasegmental variables used ($f_0$, energy envelope, duration) seem to work seamlessly to the same end, which is to signal the hierarchical and parallel structure of the linguistic signals.  The role of signal energy as a reliable determinant of prosodic structure is interesting, but not altogether surprising \cite{kochanski2005loudness}.  On the one hand, it diminishes the role of $f_0$, while on the other hand, it also provides it with more freedom for other (post-lexical) prosodic functions that are not strictly related to the hierarchical structure.

As mentioned above, the methods and representations brought forward in this study have been designed to be feasible in a broader scientific spectrum keeping in mind their psychological plausibility. Although the wavelet representation of prosody has a strong correspondence with the manual annotations of the evaluation corpus (highlighting their relationship with perception), the neural computations performed by the auditory system might differ considerably in contributing to the percepts underlying the accent and break judgments of the labellers. In particular, the scheme for iteratively filling the gaps in the acoustic signals is not a plausible algorithm for neural processing. However, the assumed temporal integration model to explain silent gap detection gives similar ``filling'' behaviour as the current processing of gain signal. Importantly, the parameters and particular formulas were only inspired by the known auditory processes but chosen based on performance on a few test sentences. In the proposed accent and boundary annotation the wavelet analysis is performed to a few one-dimensional signals. However, a neurally more plausible approach would be a truly multidimensional representation of speech signal similar  to the multi-scale visual analyses \cite{romeny2014}.  Crucially, the wavelet trees relate the accents and boundaries together phonetically hinting at a unified mechanism, in both production and perception, between the phonetic realisation of these primary concepts of prosodic phonology.

\section*{Acknowledgements}

The research leading to these results has received funding from the European Community's Seventh Framework Programme (FP7/2007--2013)
under grant agreement n$^o$ 287678 (Simple4All) and the Academy of Finland (project n$^o$ 1265610 (the MIND programme)). We would also like to thank Juraj \v{S}imko for his insight regarding this manuscript.  Special thanks go to Paavo Alku and Tuomo Raitio for the GlottHMM collaboration.

\section*{References}


\end{document}